\documentclass{article}
\usepackage{spconf,amsmath,graphicx}
\usepackage{epstopdf}
\usepackage{array}
\usepackage{graphicx}
\usepackage{multirow}
\usepackage{color, soul}
\usepackage{cite}
\usepackage{url}
\usepackage{hyperref}
\usepackage{booktabs}
\usepackage{amsmath}
\usepackage{subfigure}

\usepackage{amssymb}

\title{Exploiting modality-invariant feature for robust multimodal emotion recognition with missing modalities}
\name{Haolin Zuo$^{ 1}$, Rui Liu$^{ 1,*}$\thanks{*: Corresponding author.}\thanks{This research is funded by the High-level Talents Introduction Project of Inner Mongolia University (No. 10000-22311201/002) and the Young Scientists Fund of the National Natural Science Foundation of China (NSFC) (No. 62206136).}, Jinming Zhao$^{ 2}$, Guanglai Gao$^{ 1}$, Haizhou Li$^{ 3}$}
\address{ $^1$ Inner Mongolia University, Hohhot, China \\$^2$ Qiyuan Lab, Beijing, China \\$^3$ The Chinese University of Hong Kong, Shenzhen, China \\
\small{zuohaolin\_0613@163.com, liurui\_imu@163.com, zhaojinming@qiyuanlab.com, csggl@imu.edu.cn, haizhouli@cuhk.edu.cn}
}
\vspace{-2mm}
\begin{document}
%
\maketitle
\begin{abstract}
\vspace{-1mm}
Multimodal emotion recognition leverages complementary information across modalities to gain performance. However, we cannot guarantee that the data of all modalities are always present in practice. In the studies to predict the missing data across modalities, the inherent difference between heterogeneous modalities, namely the modality gap, presents a challenge. 
To address this, we propose to use invariant features for a missing modality imagination network (IF-MMIN) which includes two novel mechanisms: 
1) an invariant feature learning strategy that is based on the central moment discrepancy (CMD) distance under the full-modality scenario;
2) an invariant feature based imagination module (IF-IM) to alleviate the modality gap during the missing modalities prediction, thus improving the robustness of multimodal joint representation.
Comprehensive experiments on the benchmark dataset IEMOCAP demonstrate that the proposed model outperforms all baselines and invariantly improves the overall emotion recognition performance under uncertain missing-modality conditions. We release the code at: \url{https://github.com/ZhuoYulang/IF-MMIN}.
\end{abstract}
\vspace{-1mm}

\vspace{-4mm}
\section{Introduction}
\label{sec:intro}  
\vspace{-3mm}
The study of multimodal emotion recognition with missing modalities seeks to perform emotion recognition  in realistic environments \cite{missing_modality_1, missing_modality_2}, where some data could be missing due to obscured cameras, damaged microphones, etc.
The mainstream solution for the missing modality problem can be summarized in two categories: 1) missing data  generation \cite{miss_predict_1, miss_predict_3, miss_predict_4}, 2) multimodal joint representation learning \cite{joint_representation_1, joint_representation_2}.
In \cite{miss_predict_1},  an encoder-decoder network was proposed to generate high-quality missing modality images according to the available modality, In \cite{joint_representation_2},  a translation-based method with cycle consistency loss was studied to learn joint representations between modalities. In \cite{missing_modality_1}, a Missing Modality Imagination Network, or MMIN for short, was studied to learn joint representations by predicting missing modalities, which combines the above two methods.


The modality gap between heterogeneous modalities \cite{MISA, modality-gap_1, modality-gap_2} remains an issue, which adversely affects emotion recognition accuracy. The question is how to alleviate such a modality gap. While the modalities have their unique characteristics, they share the same information in the semantic space. 
The modality-invariant feature was introduced to multimodal emotion recognition with full modality data, which shows remarkable performance. Hazarika et al. \cite{MISA} proposed the shared subspace to learn potential commonalities between modalities to reduce the influence of the modality gap. Liu et al. \cite{commonality_2} proposed discrete shared spaces for capturing fine-grained representations to improve cross-modal retrieval accuracy.
All the studies suggest that the modality-invariant feature effectively bridges the modality gap. We note that there has been no related work for emotion recognition under the missing-modality conditions.

In this work, we propose a missing modality imagination network with the invariant feature (IF-MMIN). 
Specifically, we first learn the modality-invariant feature among various modalities by using a central moment discrepancy (CMD) distance \cite{CMD} based constraint training strategy. We then design the IF-MMIN neural architecture to predict the invariant features of the missing modality from the available modality.
In this way, we  fully explore the available modality to alleviate the modality gap problem in cross-modal imagination, thus, improving the robustness of  multimodal joint representation.
The experimental results, on the benchmark dataset IEMOCAP, show that the proposed method outperforms the state-of-the-art baseline models under all missing-modality conditions.

The main contributions of this work are, 
1) we propose a CMD-based distance constraint training to learn the modality-invariant feature among full modalities; 
2) we introduce invariant features into the cross-modality imagination process to reduce the impact of the modality gap and enhance the robustness of multimodal joint representation;
and 3) experimental results on various missing modalities conditions demonstrate that the proposed IF-MMIN can perform accurate emotion recognition performance in the scenario of missing modalities.

\vspace{-2mm}
\section{IF-MMIN: Methodology}
\label{sec:model}
\vspace{-2mm}
The proposed IF-MMIN scheme first employs a central moment discrepancy (CMD) distance based invariant feature learning strategy under full-modality signals to learn the modality-specific and modality-invariant features. During IF-MMIN training, the IF-IM reads these two features to learn the robust joint representation through missing modality imagination.

\vspace{-2mm}
\subsection{CMD Distance-based Invariant Feature Learning}
\vspace{-1mm}
As shown in Fig.\ref{fig:pretrain}, the pipeline of invariant feature learning includes three modules: specificity encoder, invariance encoder, and classifier.
Specificity encoder aims to extract the high-level features $(h^a, h^v, h^t)$ from the raw features $(x^a, x^v, x^t)$ of each modality to represent the modality-specific features. 
The invariance encoder takes the modality-specific features as input to extract the modality-invariant features $H$, which is concatenation by the high-level features $(H^a, H^v, H^t)$ among all modalities. 
At last, the full-connected layer-based classifier input the concatenation of $h$ and $H$ to predict the emotion category. 
After pertaining, we will adopt pretrained specificity and invariance encoders along with the proposed IF-IM module to build the IF-MMIN architecture.

\subsubsection{Specificity \& Invariance Encoders}
\label{spe_enc}
\begin{figure}[t!]
    \centering
    \centerline{\includegraphics[width=\linewidth]{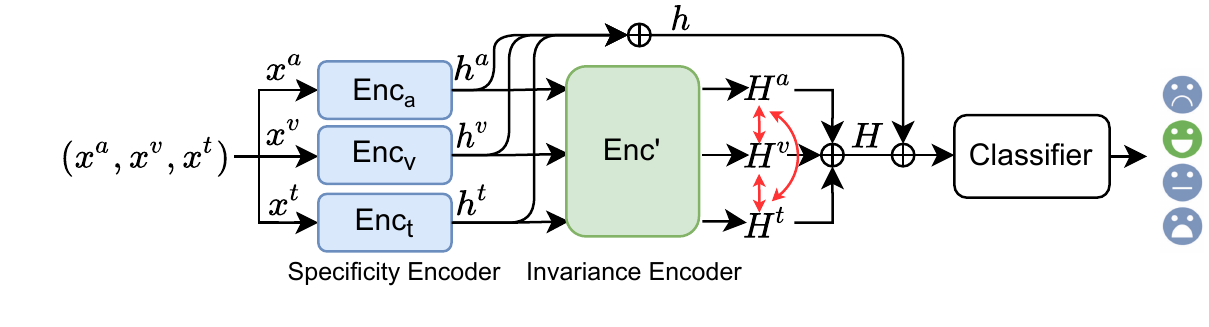}}
    \vspace{-5mm}
    \caption{The pipeline of the central moment discrepancy (CMD) distance-based invariant feature learning, which includes the specificity and invariance encoders, and classifier. The red arrows mean the CMD-based distance constraint to force various modality features to map to the same semantic subspace.
    }
    \label{fig:pretrain}
    \vspace{-4mm}
\end{figure}

As the blue blocks in Fig.\ref{fig:pretrain}, the specificity encoder is composed of three modules: acoustic, visual, and textual encoder, Enc$_{a}$, Enc$_{v}$, and Enc$_{t}$ for short respectively. 
Specifically, Enc$_{a}$ employs the LSTM \cite{LSTM} and max-pooling layer to extract the utterance-level acoustic feature $h^a$ from the raw feature $x^a$.
Enc$_{v}$ shares a similar structure with Enc$_{a}$ to read the raw features $x^v$ and output the utterance-level visual features $h^v$.
Enc$_{t}$ adopts the TextCNN \cite{TextCNN}, which is a power text representation model in NLP filed, to extract the utterance-level textual features $h^t$ from the raw feature $x^t$.

The invariance encoder, Enc$^{\prime}$, is shown as the green blocks in Fig.\ref{fig:pretrain}, which consists of the full-connected layer, the activation function, and the dropout layer. 
It aims to map modality-specific features $(h^a, h^v, h^t)$ into a shared subspace with \textit{CMD}-based distance constraint strategy (as shown by the red arrow in Fig.\ref{fig:pretrain}) to obtain high-level features $(H^a, H^v, H^t)$. 
Then, we concatenate the three high-level features into modality-invariant features $H$.


\subsubsection{CMD-based Distance Constraint}
\vspace{-2mm}
The CMD-based distance constraint aims to reduce the discrepancy between the high-level features $(H^a, H^v, H^t)$ of modalities. 
Note that CMD \cite{CMD} is a state-of-the-art distance metric that measures the discrepancy between the distribution of two features by matching their order-wise moment differences. 
We ensure that modality-invariant representation can be learned by minimizing the $\mathcal{L}_\text{cmd}$: 
\vspace{-2mm}

\begin{small}
\begin{equation}
\begin{aligned}
\mathcal{L}_{\operatorname{cmd}}&=\frac{1}{3} \!\!  \sum_{\substack{\left(m_1, m_2\right) \in \\\{(t, a),(t, v) \\(a, v)\}}} \big(  \|\mathbf{E}(H^{m_1})-\mathbf{E}(H^{m_2})\|_2 \\
&+\sum_{k=2}^K  \left\|C_k(H^{m_1})-C_k(H^{m_2})\right\|_2 \big)
\end{aligned}
\end{equation}
\end{small}
{where $\mathbf{E}(H)$ is the empirical expectation vector of the input sample $H$, and $C_k(H)=\mathbf{E}((H-\mathbf{E}(H))^{k})$ is the vector of all $k^{th}$ order sample central moments of the coordinates of $H$.}


\subsection{IF-MMIN Training}
\vspace{-2mm}

\begin{figure}[t]
    \centering
    \centerline{\includegraphics[width=1.1\linewidth]{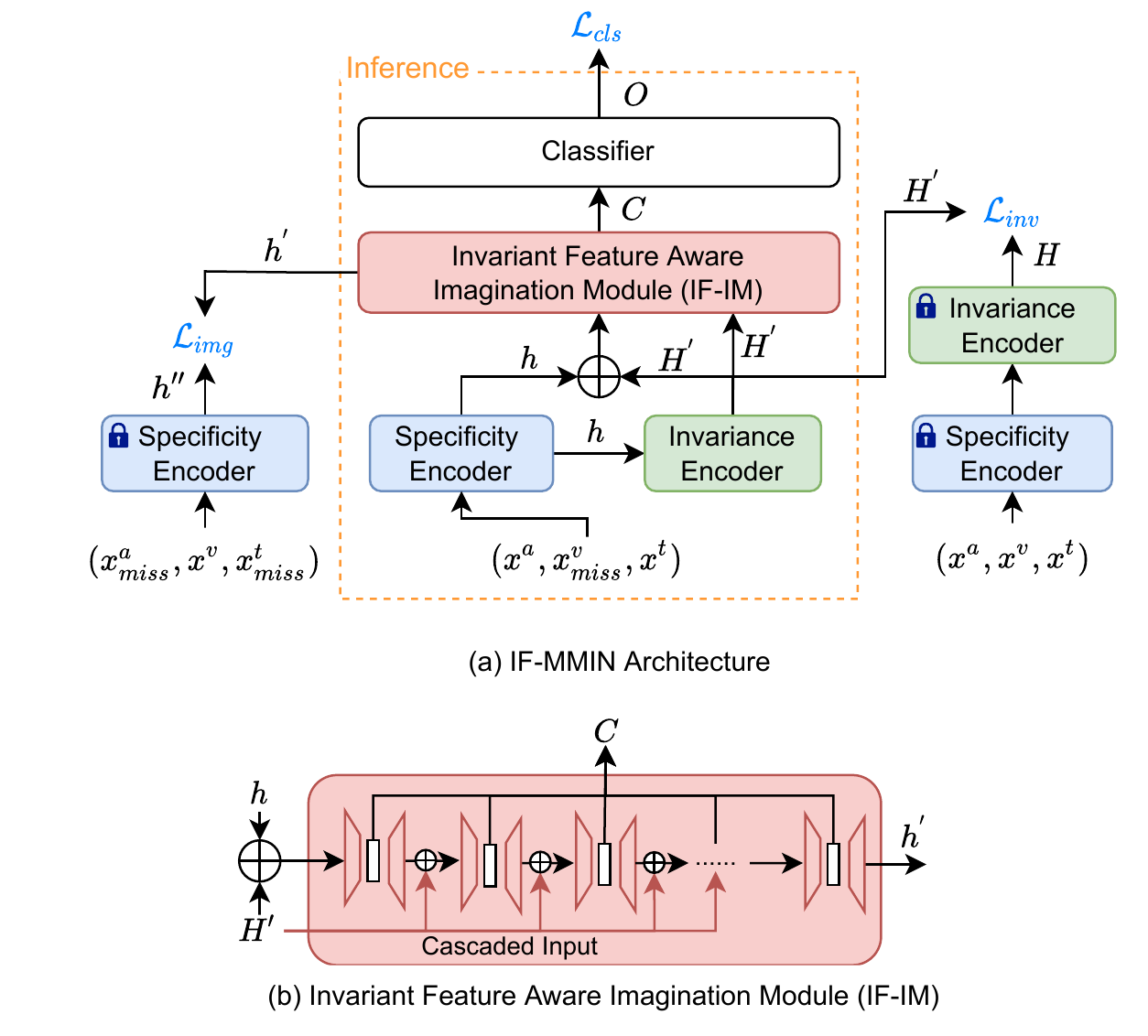}}
    \vspace{-2mm}
    \caption{The diagrams for the proposed IF-MMIN. (a) shows the overall IF-MMIN architecture, the blue locks mean their parameters are fixed during the IF-MMIN training; (b) shows the detailed structure of the modality-invariant feature aware imagination module. 
    }
    \label{fig:IF-IM}
    \vspace{-5mm}
\end{figure}

The overall architecture of IF-MMIN is illustrated in Fig.\ref{fig:IF-IM}(a), which includes 1) Specificity Encoder; 2) Invariance Encoder; 3) Modality-invariant Feature aware Imagination Module, IF-IM for short; and 4) Classifier.

Assume that the full-modalities input is $x= (x^a, x^v, x^t)$.
Specificity encoder takes $(x^a, x^v_{miss}, x^t)$,  where \textit{miss} indicates the specific missing modality, as input to extract the modality-specific features $(h^{a}, h^{v}, h^{t})$, which are then concatenated as final output $h$. 
\textcolor{black}{Invariance encoder reads $(h^{a}, h^{v}, h^{t})$ to predict the modality-invariant feature $H'$, which is a concatenation of high-level features $(H'^{a}, H'^{v}, H'^{t})$.}
$h$ and $H'$ are then fused to fed in the IF-IM to predict the feature $h'$ of the missing modality. 
The hidden features of all intermediate layers of IF-IM are then combined together as the joint representation $C$, as shown in Fig.\ref{fig:IF-IM}(b), to predict the final emotion category $O$. 
To ensure the stability of the modality-invariant feature prediction and missing modality imagination processes in IF-MMIN, \textit{Imagination Loss} ($\mathcal{L}_{img}$) and \textit{Invariance Loss} ($\mathcal{L}_{inv}$) are added on the basis of \textit{Classification Loss} ($\mathcal{L}_{cls}$).
Note that the parameters of the specificity and invariance encoders are initialized by the corresponding modules as mentioned in Sec. \ref{spe_enc}.

\vspace{-3mm}
\subsubsection{Invariant Feature Aware Imagination Module (IF-IM)}
\vspace{-1mm}
As shown in Fig.\ref{fig:IF-IM}(b), IF-IM is built with the cascaded autoencoder which includes $M$ autoencoders. Different from \cite{missing_modality_1}, IF-IM reads the $h$ and $H'$ simultaneously. In addition, $H'$ is a cascaded input given to each autoencoder to assist the missing modality imagination and alleviate the modality gap problem.

Each autoencoder denoted as $\omega_i$, i = 1,2,...,$M$. Then the calculation of each autoencoder can be defined as:

\vspace{-4mm}
\begin{equation}
    \left\{\begin{array}{l}
    \Delta z_i=\omega_i(H'+h), \quad i=1 \\
    \Delta z_i=\omega_i\left(H'+\Delta z_{i-1}\right), \quad 1<i \leq M
    \end{array}\right.
    \vspace{-2mm}
\end{equation}
where $\Delta z_i$ is the output of the $i^{th}$ autoencoder. The imagined missing modality $h'$ of IF-IM can be defined as: $h^{\prime}=\Delta z_M$.

\vspace{-3mm}
\subsubsection{Loss Functions}
During IF-MMIN training, classification loss $\mathcal{L}_{\text{cls}}$ is used to supervise the training with emotion category target $\hat O$: $ \mathcal{L}_{\text {cls }}=\text{CrossEntropy}(O, \hat O)$.
More important, \textit{Imagination Loss} $\mathcal{L}_{\text {img}}$ is used to minimize the distance between the IF-IM output $h'$ and the modality-specific feature of the missing modality $x^{v}$: $\mathcal{L}_{\text {img}}=\text{RMSE }(h_i^{\prime \prime}, h_i^{\prime})$, while \textit{Invariance Loss} $\mathcal{L}_{\text {inv}}$ aims to force the predicted modality-invariant feature $H'$ and the target modality-invariant feature $H$ of full-modality signals are close to each other: $\mathcal{L}_{\text {inv}}=\text{RMSE }(H_i, H_i^{\prime})$.



The total loss function is the sum of the three functions: $\mathcal{L}=\mathcal{L}_{\text {cls}}+\lambda_1 \mathcal{L}_{\text {img}}+\lambda_2 \mathcal{L}_{\text {inv}}$, where $\lambda_1$ and $\lambda_2$ are the balance factors.

\setlength{\tabcolsep}{0.007 \linewidth}{ 

\begin{table*}[] \scriptsize \centering
\begin{tabular}
{p{23mm}|cc|cc|cc|cc|cc|cc|ccc}
\toprule 
\multirow{3}{*}{ \qquad \quad System}   & \multicolumn{14}{c}{Testing  Conditions}  \\ \cline{2-15} 
& \multicolumn{2}{c}{\{a\}}    & \multicolumn{2}{c}{\{v\}}  & \multicolumn{2}{c}{\{t\}}    & \multicolumn{2}{c}{\{a,v\}}   & \multicolumn{2}{c}{\{a,t\}}   & \multicolumn{2}{c}{\{v,t\}}   & \multicolumn{2}{c}{Average}    \\ \cline{2-15} 
& WA                         &  {UA}                & WA                &  {UA}                & WA                         &  {UA}                         & WA                         &  {UA}                         & WA                         &  {UA}                         & WA                         &  {UA}                         & WA                         & UA                         \\ \hline
MCTN \cite{joint_representation_2}   & 0.4975  &  {0.5162}  & 0.4892  &  {0.4573} & 0.6242 &  {0.6378} & 0.5634 & {0.5584}  & 0.6834 &  {0.6946} & 0.6784 &  {0.6834} & 0.5894 & 0.5913 \\ \hline
MMIN  \cite{missing_modality_1}                                                          & 0.5511                     &  {0.5726}            & 0.5299            &  {0.5117}            & 0.6649                     &  {0.6769}                     & 0.6491                     &  {0.6584}                     & 0.7328                     &  {0.7467}                     & 0.7244                     &  {0.7307}                     & 0.6420                     & 0.6495                     \\ \hline
 MMIN w/o cycle \cite{missing_modality_1} & 0.5503                     &  0.5821           & 0.5116            &  {0.5006}            & 0.6577                     & {0.6705}                     & 0.6239                     &  {0.6454}                     & 0.7185                     & 0.7438                    & 0.7202                     &  0.7301                      & 0.6304                     & 0.6454                     \\ \hline
\textbf{IF-MMIN} \textbf{(ours)} & \textbf{0.5620 $^{\uparrow}_{\Uparrow}$ } &  {\textbf{0.5813 $^*_{\Uparrow}$ }} & \textbf{0.5197 $^*_{\Uparrow}$ } &  {\textbf{0.5041 $^*_{\Uparrow}$ }} & \textbf{0.6702 $^{\uparrow}_{\Uparrow}$ } &  {\textbf{0.6820 $^{\uparrow}_{\Uparrow}$ }} & \textbf{0.6533 $^{\uparrow}_{+}$} &  {\textbf{0.6652 $^{\uparrow}_{+}$}} & \textbf{0.7405 $^{\uparrow}_{\Uparrow}$ } &  {\textbf{0.7544 $^{\uparrow}_{\Uparrow}$ }} & \textbf{0.7268 $^{\uparrow}_{\Uparrow}$ } &  {\textbf{0.7362 $^{\uparrow}_{\Uparrow}$ }} & \textbf{0.6454 $^{\uparrow}_{\Uparrow}$ } & \textbf{0.6538 $^{\uparrow}_{\Uparrow}$ } \\ \hline
\quad w/o   $\mathcal{L}_{\text {inv}}$                              & 0.5513                     &  {0.5767}            & 0.5156            &  {0.4955}            & 0.6631                     &  {0.6768}                     & 0.6511                     &  {0.6665}                     & 0.7333                     &  {0.7511}     & 0.7163     &  {0.7277}      & 0.6385                     & 0.6490     \\ \hline
\quad w/o cascaded input           & 0.5552     &  {0.5768}            & 0.5159            &  {0.5038}            & 0.6642                     &  {0.6790}                     & 0.6536                     &  {0.6650}                     & 0.7337                     &  {0.7530}                     & 0.7173                     &  {0.7300}                     & 0.6400                     & 0.6514   \\      \bottomrule 
\end{tabular}
\vspace{-2mm}
\caption{The experimental results of our IF-MMIN, three baselines, and two ablation systems under six missing-modality conditions (i.e. testing condition $\lbrace t \rbrace$ indicates that only textual modality is available and both visual and acoustic modalities are missing.). ``Average'' means the average performance in overall conditions. $\uparrow$ and $\Uparrow$ indicate that the current result outperforms all baselines and ablation systems respectively. * and + indicate that the results are on par with those of the best baseline or best ablation system respectively.
}
\label{main results}
\vspace{-5mm}
\end{table*}
}

\section{Experiments and Results}
\label{sec:exp}
\vspace{-2.5mm}
We validate the IF-MMIN on the Interactive Emotional Dyadic Motion Capture (IEMOCAP) dataset \cite{IEMOCAP}.  
Following \cite{missing_modality_1}, we process IEMOCAP emotional labels into four categories: happy, angry, sad, and neutral. 
The splitting ratio of training/validation/testing sets is 8:1:1.
\vspace{-3mm}
\subsection{Experimental Setup}
\vspace{-2mm}
Similar to those in \cite{missing_modality_1}, the raw features $x^a$, $x^v$ and $x^t$ are 130-dim OpenSMILE \cite{Opensmile} features with the configuration of ``IS13\_ComParE'', 342-dim ``Denseface'' features extracted by a pretrained DenseNet model \cite{DenseNet} and 1024-dim BERT word embeddings, respectively.

The hidden size of specifically encoders Enc$_a$ and Enc$_v$ is set to 128, Enc$_t$ contains 3 convolution blocks with kernel sizes of {3,4,5} and the output size of 128.
The size of invariance encoder Enc$^{\prime}$ output, $\mathcal{H}$, is 128.
IF-IM consists of 5 autoencoders in size 384-256-128-64-128-256-384, where the hidden-vector size is 64.
The classifier includes 3 fully-connected layers of size \{128,128,4\}.
Since the value of $\mathcal{L}_{\text {inv}}$ is quite smaller (about 1\%) than $\mathcal{L}_{\text {img}}$, we set $\lambda_1$ is 1 and $\lambda_2$ is 100 to balance the numerical difference and elevate the importance of $\mathcal{L}_{\text {inv}}$ in the total loss. The batch size is 128 and the dropout rate is 0.5.
We adopt the Adam optimizer \cite{DBLP:journals/corr/KingmaB14} which with the dynamic learning rate and the initial rate is 0.0002, and use the Lambda LR \cite{wu2020deep} to update the learning rate.

We conduct all experiments, including invariant feature learning and IF-MMIN training, with 10-fold cross-validation, where each fold contains 40 epochs. 
To demonstrate the robustness of our models, we run each model three times to alleviate the influences of random initialization of parameters.
We select the best model on the validation set and report its performance on the testing set.
All models are implemented with Pytorch deep learning toolkit and run on a single NVIDIA Tesla P100 graphic card.

\vspace{-3mm}
\subsection{Comparative Study}
\vspace{-1mm}
We develop three multimodal emotion recognition systems for a comparative study.
1) \textbf{\textit{MCTN} }\cite{joint_representation_2} learns the joint representation via a cyclic translation between missing and available modalities;
2) \textbf{\textit{MMIN}} \cite{missing_modality_1} is the state-of-the-art model for the missing modality problem which learns the joint representation through cross-modality imagination via the autoencoder and cycle consistency learning; 3) \textbf{\textit{MMIN w/o cycle}} \cite{missing_modality_1} removes the cycle consistency learning part of MMIN and just retains the forward missing modality imagination process, which is a fair counterpart for our IF-MMIN.

\vspace{-3mm}
\subsection{Main Results for Uncertain Missing-Modality}
\vspace{-1mm}
To validate our IF-MMIN under different missing modality testing conditions \cite{missing_modality_1}, 
We report all results in terms of \textit{Weighted Accuracy} (WA) \cite{WA} and \textit{Unweighted Accuracy} (UA) \cite{UA}.

As shown in rows 2 through 5 of Table \ref{main results}, our IF-MMIN achieves the highest average values under all missing-modality testing conditions. For each condition, IF-MMIN also outperforms all baselines except for conditions \{a\} and \{v\}, where it is comparable to the optimal baseline. The possible reason is that the textual modality contains more semantic information than the audio and visual modalities \cite{DBLP:conf/icmcs/FuLXQFZL22}.
In a nutshell, all the results show that IF-MMIN can learn robust multimodal joint representation, alleviate the modality gap by introducing modality-invariant features, and thus enable remarkable performance under different missing-modality testing conditions.

\vspace{-4mm}
\subsection{Ablation Study}
\vspace{-1mm}
IF-MMIN exploits the invariant feature $H'$ and adds the Invariance Loss $\mathcal{L}_{inv}$ to boost the missing modality imagination and IF-IM takes the invariant feature $H'$ with cascaded input. To verify their contributions, we conduct the following ablation experiments: 
1) \textbf{IF-MMIN w/o $\mathcal{L}_{\text {inv}}$} system discards the  $\mathcal{L}_{\text {inv}}$ during IF-MMIN training. 2) \textbf{IF-MMIN w/o cascaded input} system just only takes $H'$ as the input of the first autoencoder, instead of each autoencoder, in IF-IM.
 
As shown in rows 5 through 7 of Table \ref{main results}, \textit{IF-MMIN} also outperforms the \textit{IF-MMIN w/o $\mathcal{L}_{\text {inv}}$} and \textit{IF-MMIN w/o cascaded input} in most cases, which confirms that 1) the invariance encoder of IF-MMIN can predict accurate invariant feature under the constraints of $\mathcal{L}_{\text {inv}}$, so as to better serve the IF-IM; 2) the cascaded input can provide prior knowledge when each layer of autoencoder works and indeed strengthen the imagination ability of IF-IM.

\vspace{-3mm}
\subsection{Visualization Analysis}
\vspace{-2mm}
The accuracy of invariant feature learning is the premise for IF-MMIN to work well. Therefore, to verify the role of invariant feature learning related modules, including $\mathcal{L}_{cmd}$, $\mathcal{L}_{inv}$, $H'$ and $H$, we conduct the following visualization experiments for IF-MMIN.

\begin{figure}
\vspace{-4mm}
    \subfigure[Invariant features visualization]{
        \includegraphics[width=0.45\linewidth]{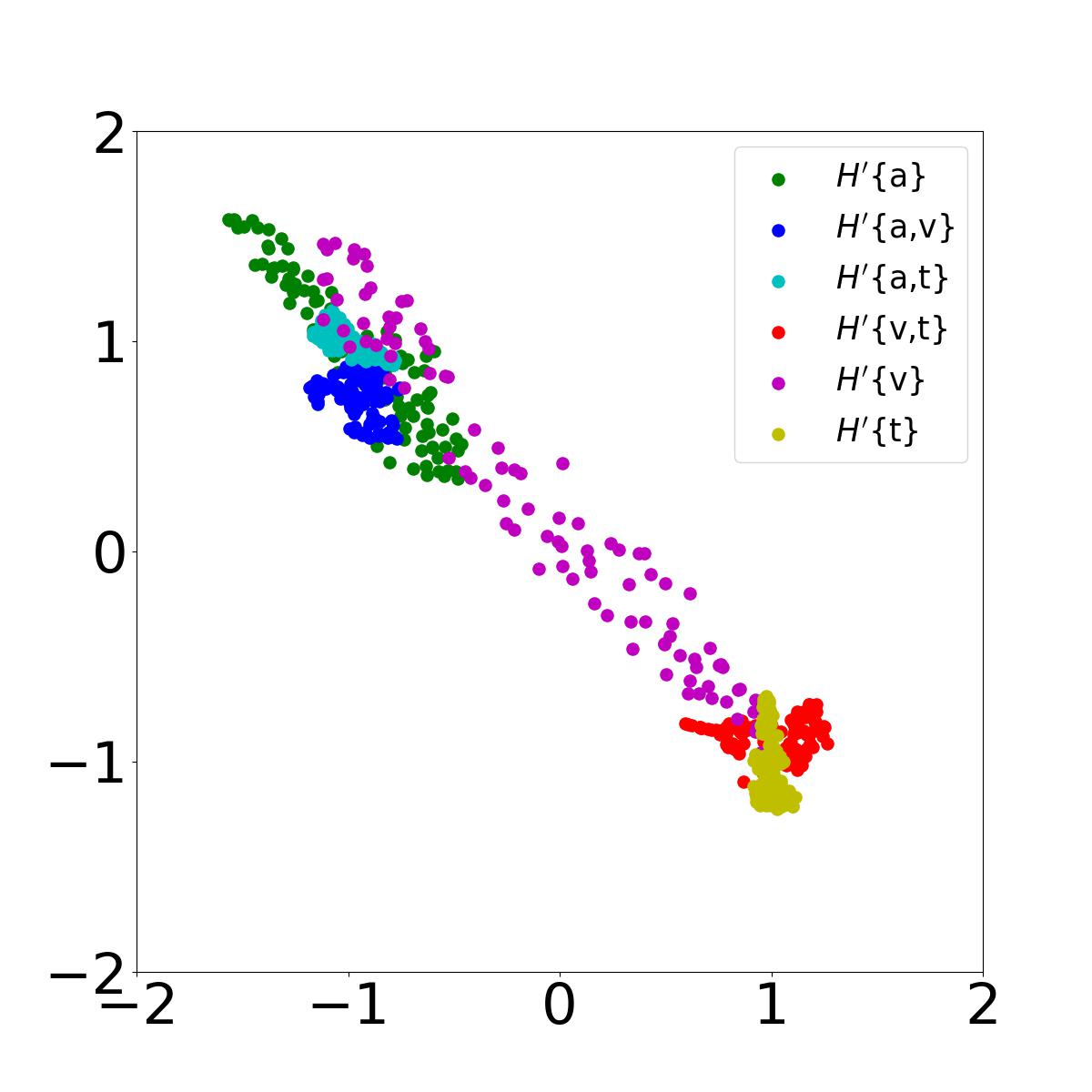}
            
    }
    \subfigure[Convergence trajectory of $\mathcal{L}_{\text {inv}}$]{    
        \includegraphics[width=0.45\linewidth]{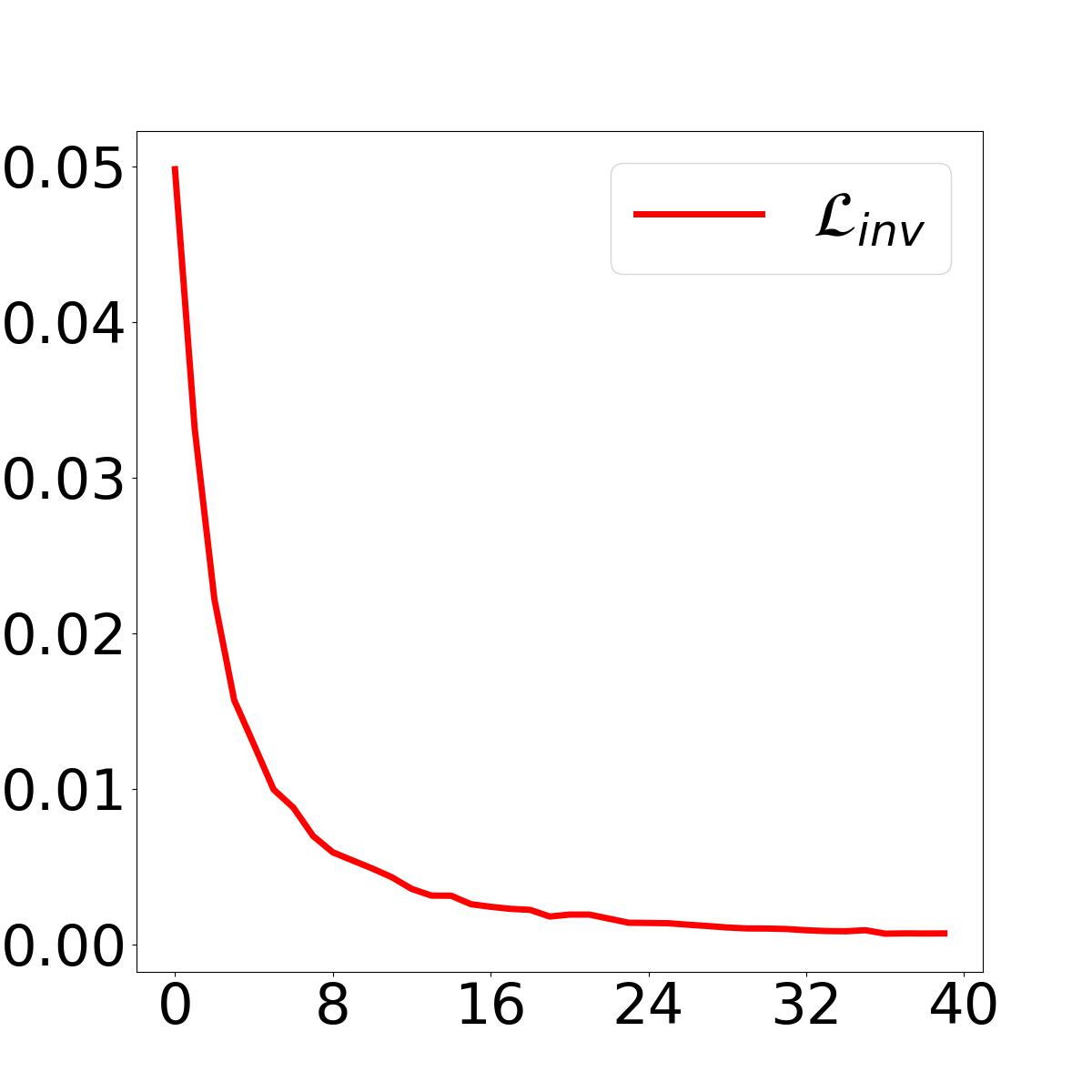}
    }
    \vspace{-3mm}
    \caption{Visualization analysis about the invariant feature and the $\mathcal{L}_{\text {inv}}$. (a) 
    shows the t-SNE plot of the distribution of the predicted  modality-invariant features $H'$ under all six missing-modality conditions. (b) denotes the convergence trajectory of $\mathcal{L}_{\text {inv}}$ during the IF-MMIN training. (the x-axis represents the iteration number and the y-axis represents the loss value).}
    \vspace{-4mm}
    \label{invariance and loss}
\end{figure}

We visualize the $H'$ under six missing conditions using t-SNE algorithm in a two dimensional plane \cite{van2008visualizing}, as shown in Fig. \ref{invariance and loss}(a). 
We randomly select 600 sentences from the testing set, 100 sentences for each condition, and extract 600 invariant features $H'$. Therefore, there are 600 points in Fig. \ref{invariance and loss}(a), 100 points in each color.
It's observed that all $H'$ under various conditions form a clear clustering in terms of feature distributions, which is encouraging.
Fig. \ref{invariance and loss}(b) shows the convergence
the trajectory of $\mathcal{L}_{inv}$ during the IF-MMIN training, where the x-axis represents the epoch and the y-axis represents the loss value. The smooth curve in the figure shows that $H'$ and $H$ are close to each other in the training process, thus further proving the effectiveness of $\mathcal{L}_{inv}$. 
Since $H$ is learned under the constraint of $\mathcal{L}_{cmd}$, it also proves the effectiveness of $\mathcal{L}_{cmd}$.


\vspace{-3mm}
\section{Conclusion}
\label{sec:con}
\vspace{-2mm}
This work studied a novel invariant feature aware multimodal emotion recognition model (IF-MMIN) that includes a CMD-based distance-based invariant feature learning and the invariant feature aware missing modality imagination module (IF-IM). By exploiting the invariant feature, our IF-MMIN can alleviate the modality gap and improve the robustness of the multimodal joint representation.
Experimental results on IEMOCAP and demonstrated that the proposed IF-MMIN outperforms the start-of-the-art baselines under various missing modality conditions. In future work, we will explore ways to further improve invariant feature learning.
 %

 

 
\bibliographystyle{IEEEbib}
{\footnotesize
\bibliography{strings}}

\end{document}